\documentclass[a4paper, 10pt, conference]{ieeeconf}      % Use this line for a4
\usepackage{FG2021}
\usepackage{graphicx}
\usepackage[]{algorithm2e}
\usepackage{subcaption}
\usepackage{multirow}
\usepackage{float}
\usepackage[pagebackref=true,breaklinks=true,letterpaper=true,colorlinks,bookmarks=false]{hyperref}
\usepackage{caption}
\usepackage{amssymb}
\usepackage{amsmath}

\FGfinalcopy % *** Uncomment this line for the final submission

\IEEEoverridecommandlockouts                              % This command is only
\def\FGPaperID{307} % *** Enter the FG2021 Paper ID here

\title{\LARGE \bf
Face Trees for Expression Recognition
}

\author{\parbox{16cm}{\centering
    {\large Mojtaba Kolahdouzi, Alireza Sepas-Moghaddam, and Ali Etemad}\\
    {\normalsize
    Dept. ECE and Ingenuity Labs Research Institute, Queen’s University, Kingston, Canada\\
    }}
    \thanks{This work was funded by Irdeto Canada Corporation and the Natural Sciences and Engineering Research Council of Canada (NSERC).}% <-this % stops a space
    \thanks{\copyright{} 2021 IEEE.  Personal use of this material is permitted.  Permission from IEEE must be obtained for all other uses, in any current or future media, including reprinting/republishing this material for advertising or promotional purposes, creating new collective works, for resale or redistribution to servers or lists, or reuse of any copyrighted component of this work in other works}
}

\begin{document}

% Case #4: For all other papers the copyright notice is:
% \IEEEoverridecommandlockouts\pubid{\makebox[\columnwidth]{978-1-6654-3176-7/21/\$31.00~\copyright{}2021 IEEE \hfill}
% \hspace{\columnsep}\makebox[\columnwidth]{ }}

\ifFGfinal
\thispagestyle{empty}
\pagestyle{empty}
\else
\author{Anonymous FG2021 submission\\ Paper ID \FGPaperID \\}
\pagestyle{plain}
\fi
\maketitle

%%%%%%%%%%%%%%%%%%%%%%%%%%%%%%%%%%%%%%%%%%%%%%%%%%%%%%%%%%%%%%%%%%%%%%%%%%%%%%%%
\begin{abstract}
We propose an end-to-end architecture for facial expression recognition. Our model learns an optimal tree topology for facial landmarks, whose traversal generates a sequence from which we obtain an embedding to feed a sequential learner. The proposed architecture incorporates two main streams, one focusing on landmark positions to learn the \textit{structure} of the face, while the other focuses on patches around the landmarks to learn \textit{texture} information. Each stream is followed by an attention mechanism and the outputs are fed to a two-stream fusion component to perform the final classification. We conduct extensive experiments on two large-scale publicly available facial expression datasets, AffectNet and FER2013, to evaluate the efficacy of our approach. Our method outperforms other solutions in the area and sets new state-of-the-art expression recognition rates on these datasets.

\end{abstract}

%%%%%%%%%%%%%%%%%%%%%%%%%%%%%%%%%%%%%%%%%%%%%%%%%%%%%%%%%%%%%%%%%%%%%%%%%%%%%%%%
\section{Introduction}

Facial expressions have been regarded as one of the most common non-verbal cues for conveying human emotions~\cite{surveyAsli, wang2020suppressing}. Therefore, automatic facial expression recognition (FER) from images has received considerable attention in recent years~\cite{sepas2021capsfield, sepas2020facial, roy2021spatiotemporal}. Naturally, deep learning methods have recently become the dominant approach for FER, starting with early work using a shallow CNN and a support vector machine (SVM) in~\cite{tang2013deep}. More sophisticated deep learning solutions have recently been proposed. In \cite{chen2019facial}, a novel attention method which causes the method to focus on moving facial muscles was proposed. In ~\cite{wang2020region} another attention-based method was proposed to rank the importance of facial regions. To tackle data imbalance, which is a common issue in FER, a novel loss function was proposed in~\cite{jiang2020accurate}. In~\cite{kollias2020deep}, a novel data augmentation approach for FER was proposed. By altering the standard skip connections in a ResNet model with a differentiable function, the Bounded Residual Gradient Network (BReG-Net) was proposed in~\cite{hasani2019bounded}. By adding trainable parameters to BReG-Net, the same authors later proposed BReG-NeXt~\cite{hasani2020breg}, which offers state-of-the-art results on AffectNet~\cite{Affectnet} and FER2013~\cite{FER2013} datasets. More recently, self-supervised and contrastive approaches have also begun to gain popularity in this area, but have in many cases used either videos~\cite{roy2021spatiotemporal} or multi-view~\cite{shuvendu2021} data as opposed to the work presented in this study which focuses on a single image.

\begin{figure}[!t]
    \begin{center}
    \includegraphics[width=0.9\columnwidth]{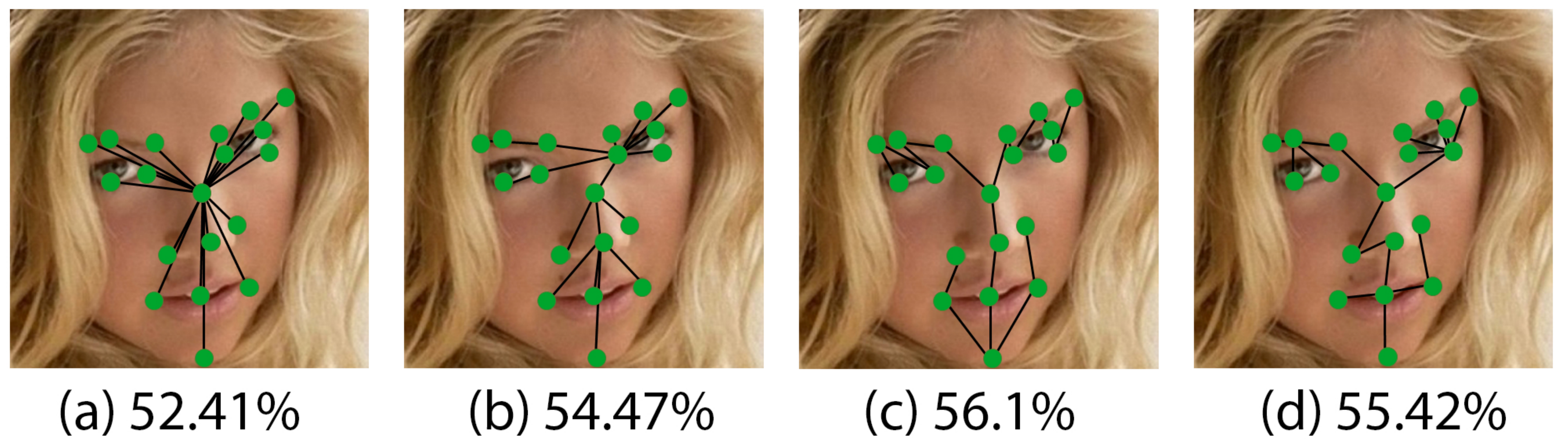}
    \end{center}
\caption{A face image along with 4 random trees. Corresponding recognition rates are written below each image.}
\label{fig:banner}
\end{figure}

In order to make use of the existing \textit{relationships} between facial regions, exploiting graph topologies in neural networks for FER has gained popularity~\cite{GCN1,G1,ICLR}. In~\cite{GRNN}, Gabor features were combined and used to feed a gated recurrent unit. In~\cite{G1}, the relationship between facial landmarks were provided by using their psychological semantic relationships. The resulting topology was then processed by a graph network. Despite clear effectiveness of such graph-based methods, \textbf{there is no unique consensus on the topology of the underlying graphs} that should be used for FER. Current graph-based FER methods either construct complete graphs~\cite{GRNN, GCN2,G2} or use pre-defined topologies~\cite{G1,G3,GCN1}, both of which do not capture the optimal connections. The topologies of these graphs can have considerable effects on the performance of networks, including sequential learners. Figure~\ref{fig:banner} shows a face image along with 4 random trees that were built on the extracted facial landmarks. Traversing these trees results in 4 different sequence-based embeddings that can be fed to 4 independent LSTM networks for a 7-class FER task. The corresponding recognition rates are shown below each image in the figure where we see a considerable range given the different topologies.

In this paper, we propose an end-to-end deep FER architecture to learn an optimal graph of the face in the form of a tree topology whose traversal creates a sequence, which is used to form the input embedding for a sequence learner. To this end, we consider a weighted complete graph, from which we generate a minimum-cost spanning tree and choose the optimal edges. Our end-to-end model 
% named face topology network (FaceTopoNet) 
consists of three blocks. In the first block, we perform the tree topology learning. Our optimized tree is then used in a two-stream framework, structure and texture. In the structure stream, we traverse the optimized tree to form a sequence, whose resultant embedding is then fed to an LSTM. This is followed by a soft attention mechanism. In the texture stream, we encode the local patches around the extracted facial landmarks, prior to learning a separate LSTM network. The final embedding is fused with that of the structure stream. To the best of our knowledge, this paper is the first that optimizes the order of relevant facial regions prior to recurrently learning their dependencies for FER. Our experiments on two large-scale public FER datasets, AffectNet~\cite{Affectnet} and FER2013~\cite{FER2013}, set new state-of-the-art results.

In summary, our main contributions are as follows. (\textbf{1}) We propose a novel end-to-end architecture to utilize optimal face tree topologies. (\textbf{2}) Our experiments show that the two-stream solution has a positive impact on the results. Moreover, we observe that given our tree topology optimization, a different face tree is learned for each dataset. Lastly, our solution sets new \textbf{state-of-the-art} values on large-scale AffectNet~\cite{Affectnet} and FER2013~\cite{FER2013} datasets.

%%%%%%%%%%%%%%%%%%%%%%%%%%%%%%%%%%%%%%%%%%%%%%%%%%%%%%%%%%%%%%%%%%%%%%%%%%%%%%%%
\section{Proposed Method}

\subsection{Problem and Solution Overview}
\begin{figure}
    \begin{center}
    \includegraphics[width=0.7\columnwidth]{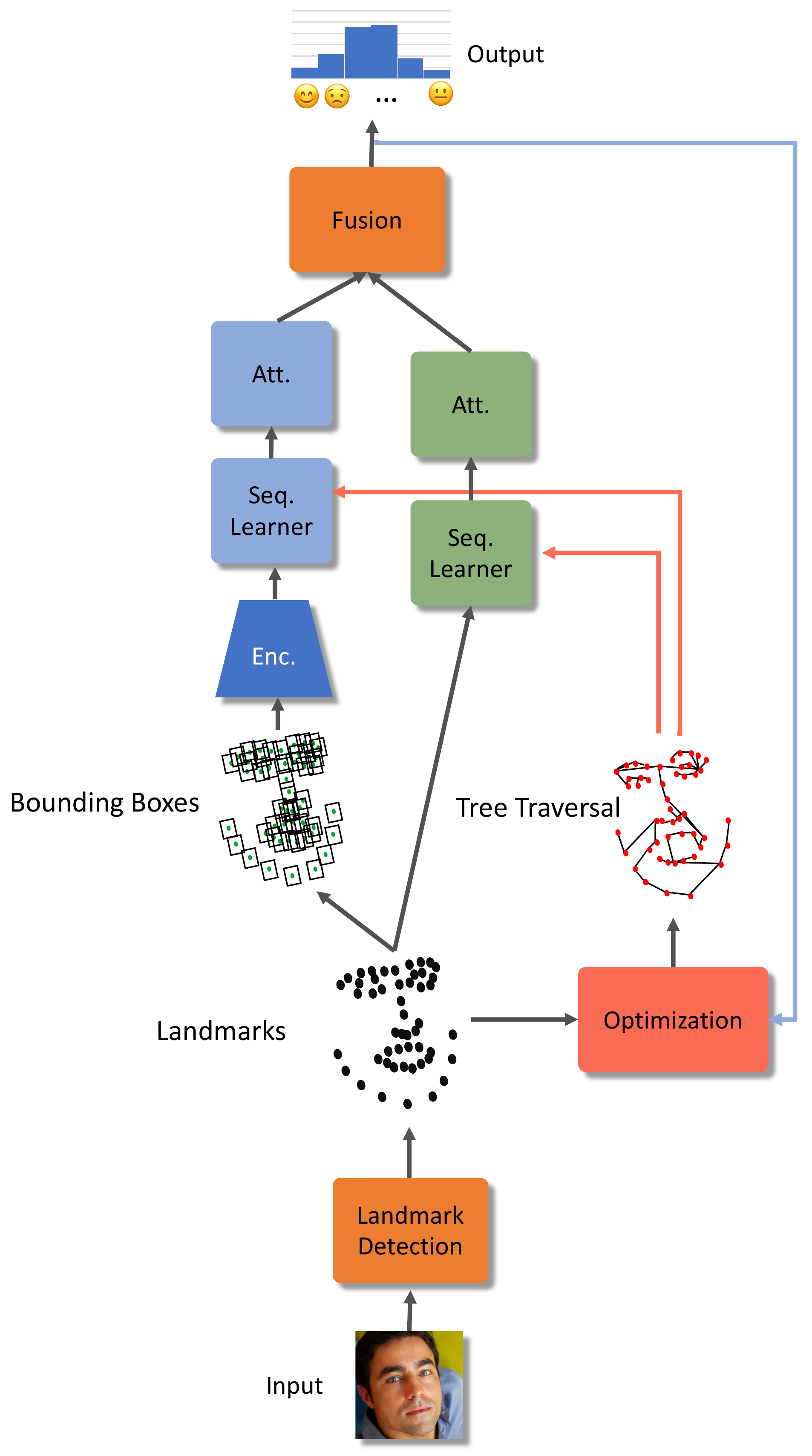}
    \end{center}
% \caption{Our proposed FaceTopoNet pipeline.}
\caption{Our proposed pipeline.}
\label{fig:pipeline}
\end{figure}

Graphs can model facial components~\cite{G1, angadi2019face} in which vertices denote facial landmarks while edges correspond to the relationship between these landmarks~\cite{jiang2017emotion}. As evident from Figure~\ref{fig:banner}, the topology of such graphs can affect the performance of subsequent models that use it. Nevertheless, no prior work has been proposed to optimally learn the topology of facial graphs. To address this gap, we propose a method to learn the optimum tree capable of characterizing these relationships. Next, in order to learn critical information from the face, we consider two key streams, \textit{structure} and \textit{texture}. The former learns the structure while the latter learns the texture of key patches in the face, both of which consider the optimized tree. Our method traverses the optimized tree to form a sequence, which is then learned recurrently using a sequential learner, in this case an LSTM. Figure~\ref{fig:pipeline} illustrates the architecture of our proposed pipeline.

\subsection{Tree Topology Learning}
\begin{figure}
    \begin{center}
    \includegraphics[width=1\columnwidth]{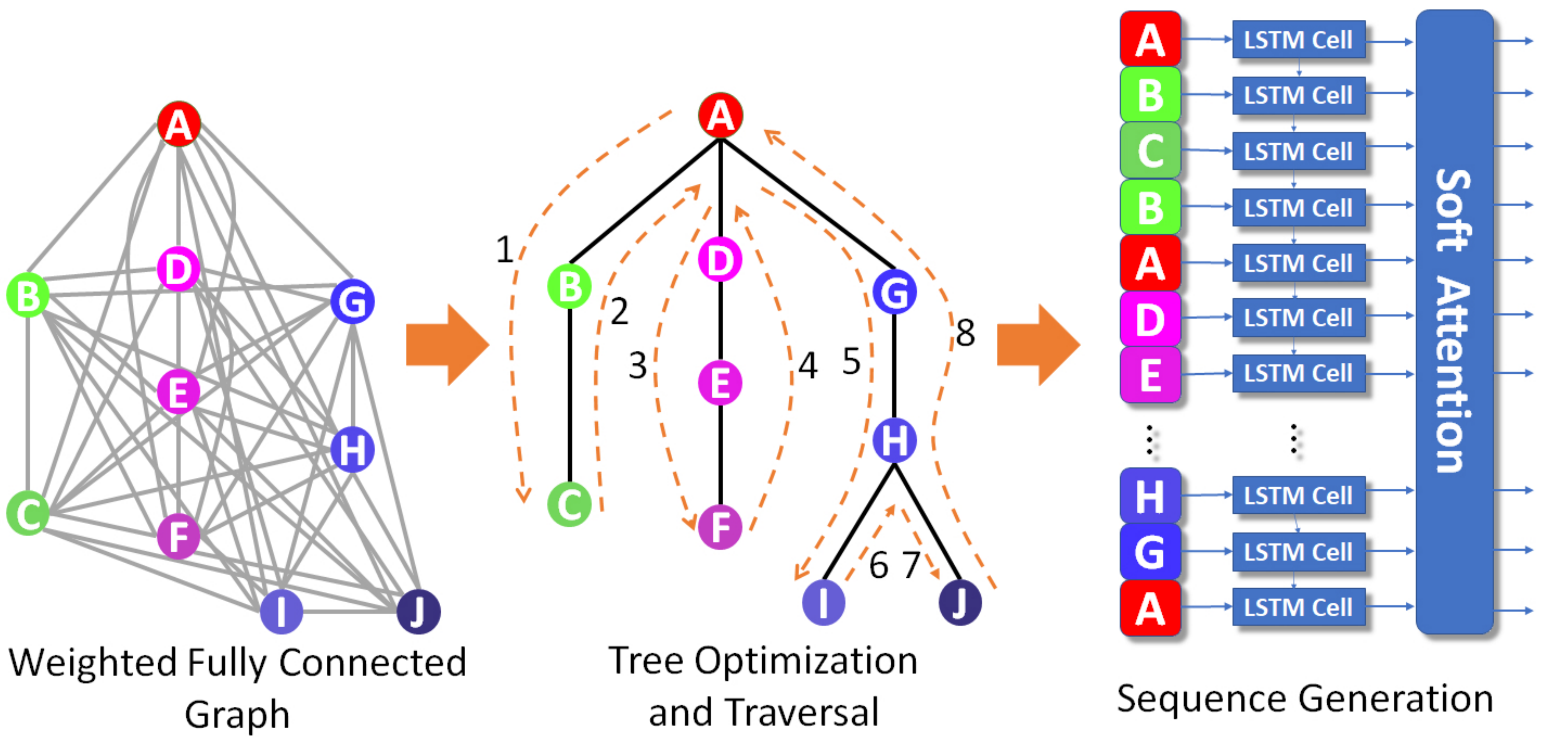}
    \end{center}
\caption{Details of the face tree topology learning. Traversing the tree results in A-B-C-B-A-D-E-F-E-D-A-G-H-I-H-J-H-G-A, which is then fed to an LSTM network.}
\label{fig:TreeTraversal}
\end{figure}

We intend to first learn an optimal face tree topology to be used in subsequent components. To this end, we first detect the facial landmarks from the input images using a deep regression architecture~\cite{lv2017deep}. Next, we construct a weighted fully connected graph, $k_n$, from the landmarks. We then find the optimum minimum-cost spanning tree considering the weights of $k_n$, whose traversal generates a sequence by which we form an embedding to feed to a sequential learner. We use the Prim's algorithm~\cite{jiang2009research} to solve the problem of finding a minimum-cost spanning tree. The built tree is then traversed using Preorder depth-first order~\cite{kozen1992depth}, with the starting point being the most centered landmark, i.e., nose tip. During the traversal, when a leaf node is reached, the traversal backtracks to go back to the starting point. 
% The traversal forms a sequence, which in turn is used to form an embedding as input to LSTM networks in the two subsequent structure and texture streams. 
Figure~\ref{fig:TreeTraversal} depicts the tree traversal process.

Our model then updates the learnable weights associated with $k_n$. To this end, we adopt a metaheuristic optimization algorithm as used in~\cite{li2011cooperatively}. For the optimizer's objective function we use the cost function as follows:
\begin{multline}\label{Eq:loss}
J(w)=\frac{1}{3m}(\sum\limits_{i=1}^m L^{foc}_1(\hat{y}^{(i)}, y^i) + \sum\limits_{i=1}^m L^{foc}_2(\hat{y}^{(i)}, y^i ) +\\ \sum\limits_{i=1}^m L^{foc}_3(\hat{y}^{(i)}, y^i)),
\end{multline}
where $m$ is the size of the training set, $L^{foc}_1(x)$ is the focal loss function selected to maximize performance for the fusion step, and $L^{foc}_2(x)$ and $L^{foc}_3(x)$ are respectively the focal loss functions for structure and texture streams. We use focal losses since the data distribution in most FER datasets is highly skewed~\cite{lin2017focal}. Here, $L^{foc}$ is defined as:
\begin{equation}\label{Eq:focal}
L^{foc}(p_b) = -\alpha_b(1-p_b)^\gamma log(p_b),
\end{equation}
where $\gamma$ is the focusing parameter and $p_b$ is the binomial distribution, in which $\hat{y}$ is the probability of $y$ being 1 and $1-\hat{y}$ is the probability of $y$ being 0. We set $\gamma=2$ and $\alpha_b = 0.25$ as recommended in~\cite{lin2017focal}.

At each epoch, the metaheuristic algorithm generates a set of weights associated with $k_n$, from which we generate the minimum-cost spanning tree, whose traversal results in a sequence. Next, this sequence is frozen and the subsequent structure and texture streams are fully trained. The reason for converting the obtained graph to a tree is that the acyclic nature of trees is better suited for traversal, hence sequence generation. 

\subsection{Structure Stream}
The coordinates of the facial landmarks which were extracted in the tree topology learning step, along with the obtained sequence, are the inputs to the structure stream (see Figure~\ref{fig:pipeline}). By using this sequence, we generate an embedding to be fed to an LSTM network with peephole connections~\cite{gers2000recurrent}. Following the LSTM network, we add an attention step~\cite{rocktaschel2015reasoning} to better focus on important parts. To this end, the hidden state vectors $h_i$ of the LSTM cells are multiplied by a set of coefficients $\alpha_i$ to determine the level of attention:
\begin{equation}\label{Eq:attentionWeights}
\alpha_i = \frac{e^{u_i}}{\sum_{k=1}^n e^{u_k}},
\end{equation}
where, $n$ is the number of LSTM units and $u_i$ is calculated as:
\begin{equation}\label{Eq:attentionUi}
u_i = tanh(W_hh_i+b_h).
\end{equation}
In Eq.~\ref{Eq:attentionUi}, $W_h$ and $b_h$ are the trainable weights and biases. The attentive output is calculated $H = \sum_i\alpha_i h_i$, where $H$ is the final embedding of the structure stream.

\subsection{Texture Stream}
Similar to the structure stream, the inputs to the texture stream are the facial landmarks and the sequence generated in the face tree topology learning step. However, unlike the structure stream, we focus on the texture of the facial images. To this end, we first form $n\times n$ patches centered around each facial landmark. We then crop these patches from the images and feed them to a an encoder (ResNet50) pre-trained on the VGG-Face2 dataset~\cite{cao2018vggface2}. Through empirical experimentation, we find the patch size of 17$\times$17 pixels to yield the best results. We stack these embeddings in accordance to the sequence obtained from the tree topology learning step (see Figure~\ref{fig:TreeTraversal}). The resulting embedding is then passed onto an LSTM network followed by a soft attention mechanism.

\subsection{Fusion}
In order to fuse the outputs of the structure and texture streams, we adopt the two-stream fusion strategy used in~\cite{gu2018hybrid, zhang2020rfnet}. This fusion strategy has yielded state-of-the-art results in a number of areas~\cite{gu2018hybrid}. Here, we use two dense layers as encoders for each of the input streams, and generate stream-specific features. Then, soft attention is adopted to learn the weight $\eta$ by using
% \begin{equation}\label{Eq:fusion1}
$\eta=softmax(tanh(W_f[T^*,S^*]+b_f))$,
% \end{equation}
where $W_f$ and $b_f$ are trainable weights and biases, respectively. $T^*$ and $S^*$ are the stream-specific features for texture and structure. Next, a dense layer is utilized to learn the optimal association across the embedding-specific features.
% as:
% \begin{equation}\label{Eq:fusion2}
% y=tanh(W_y[(1+\eta_T)T, (1+\eta_S)S]+b_y),
% \end{equation}
% where $y$ is the final embedding, $W_y$ and $b_y$ are trainable weights and biases, and $T$ and $S$ are the output embeddings of texture and structure streams, respectively. 
%The final classification output determined by using a softmax layer with $y$ as its input. 
% Figure~\ref{fig:fusion} illustrates the overview of our fusion strategy.

% \begin{figure}
%     \begin{center}
%     \includegraphics[width=0.75\columnwidth]{fig11.pdf}
%     \end{center}
% \caption{The fusion strategy used in FaceTopoNet.}
% \label{fig:fusion}
% \end{figure}

% \subsection{Implementation Details}
% The entire architecture has been implemented using TensorFlow~\cite{abadi2016tensorflow} and has been trained using a pair of Nvidia RTX 2080 Ti GPUs. For optimizing the weights of the two streams, we adopt ADAM optimizer~\cite{kingma2014adam} with the learning rate, first-momentum decay, and second-momentum decay of 0.001, 0.9, and 0.99, respectively. All of the reported results were obtained on the validation sets of the respective datasets similar to~\cite{hasani2020breg, kollias2020deep, vielzeuf2017temporal}. 

\section{Experiments}
\noindent \textbf{Implementation.} The entire architecture is implemented using TensorFlow~\cite{abadi2016tensorflow} and is trained using a pair of Nvidia RTX 2080 Ti GPUs. For optimizing the weights of the model, we adopt ADAM optimizer~\cite{kingma2014adam} with the learning rate, first-momentum decay, and second-momentum decay of 0.001, 0.9, and 0.99, respectively.

% In this section, we provide an overview of the two popular in-the-wild facial expression datasets which we used to evaluate our proposed method. We then outline the details of our experiments and their results.

\noindent \textbf{Datasets.} We test our model on two large in-the-wild datasets: (\textit{\textbf{i}}) \textit{\textbf{AffectNet}}, comprising more than 1 million images with 8 different expression categories, and (\textit{\textbf{ii}}) \textit{\textbf{FER2013}}, which includes 33000 images and has 6 expression categories.

% \subsection{Datasets}
% \textbf{AffectNet~\cite{Affectnet}.} This dataset was introduced in 2017, and comprises more than 1 million images, all of which have been obtained from the internet by leveraging 3 search engines. About half of the dataset has been manually annotated categorically for the presence of 7 facial expressions (anger, disgust, fear, happy, sad, surprise, and contempt) plus neutral, and dimensionally for valence and arousal intensities. The validation set of this dataset includes 500 images for each expression class.

% \textbf{FER2013~\cite{FER2013}.} This dataset was introduced in 2013 and includes 28709, 3589, and 3589 images for training, validation, and testing. All the images have been resized to 48$\times$48 pixels. Face images in this dataset are labeled as having one of the 6 basic expressions (anger, disgust, fear, happy, sad, and surprise) along with neutral.

% \begin{figure}
%     \begin{center}
%     \includegraphics[width=0.7\columnwidth]{fig5-8.pdf}
%     \end{center}
% \caption{Distribution of expressions in AffectNet and FER2013 datasets.}
% \label{fig:PMF}
% \end{figure}

\subsection{Performance}
The recognition rates (RR) obtained by 
% FaceTopoNet 
our model and other state-of-the-art benchmarking methods are presented in Table~\ref{table:results}. For AffectNet dataset, 
% FaceTopoNet 
our model shows an improvement of 1.52\% and 3.28\% over the BreG-NeXt50 and BreG-NeXt32, the current state-of-the-art FER methods. 
% It is worth mentioning that these methods have also utilized the same focal loss as their loss function, so there is a fair comparison. 
For FER2013 dataset, 
% FaceTopoNet 
our proposed pipeline outperforms the current state-of-the-art FER method, i.e., BreG-Next50, showing a performance improvement of 1.13\%.

% Table~\ref{table:results} also provides the obtained precision and recall values for our method along with the state-of-the-art solutions for AffectNet and FER2013 datasets. Additionally, the expression class breakdowns are presented. Some related works have not provided their classification breakdowns  or precision and recall values. From Table~\ref{table:results}, it is evident that FaceTopoNet achieves the highest precision and recall values for both datasets for most of the expression classes. Specifically, FaceTopoNet achieves the highest precision and recall in contempt and disgust expressions whose recognition is considered to be more difficult due to their small number of samples. Moreover, FaceTopoNet achieves the highest recall in FER2013 in 6 expressions out of 7 expressions. Additionally, FaceTopoNet achieves the highest precision values in most of the expressions for the same dataset.

\begin{figure}[t]
    \begin{center}
    \includegraphics[scale=0.12]{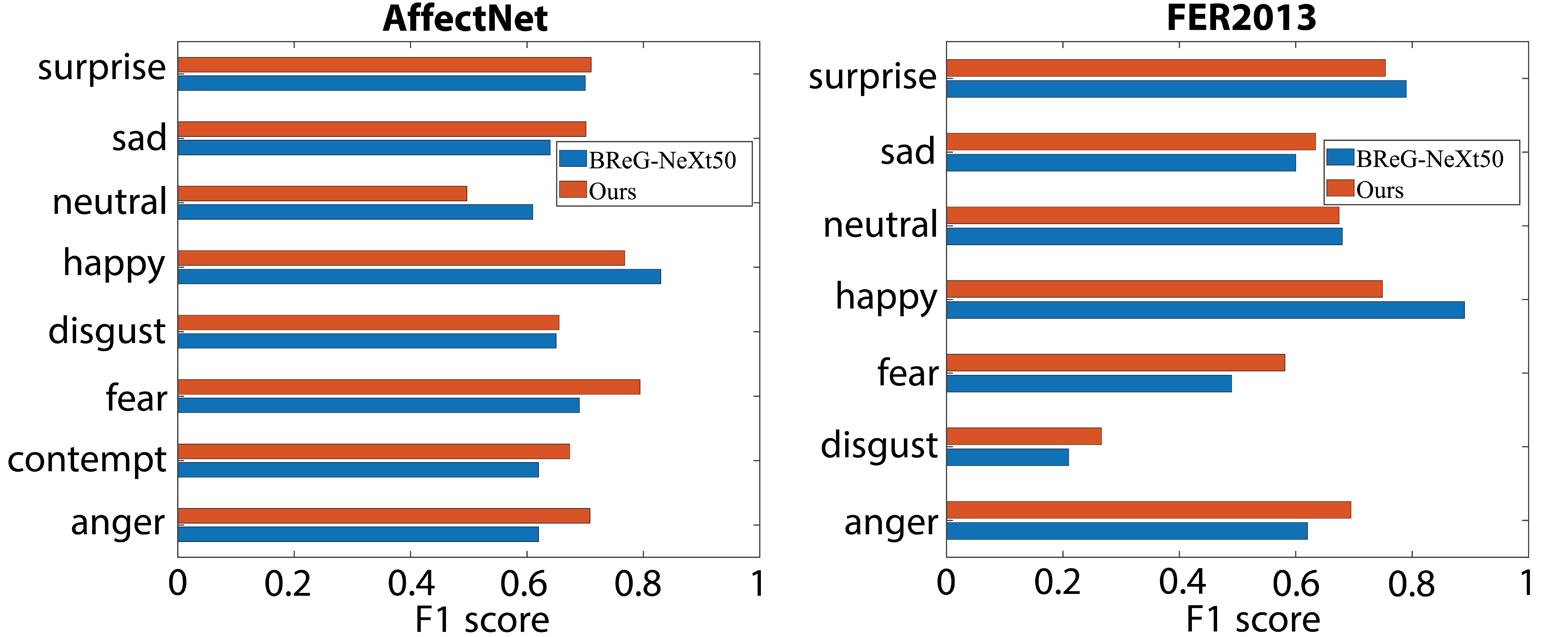}
    \end{center}
\caption{Comparison of the obtained F1 scores.}
\label{fig:f1score}
\end{figure}

For a head-to-head comparison between 
% FaceTopoNet 
our proposed model and BreG-NeXt50 (best performing benchmark), we present F1 scores for each expression by using a one-vs-all scheme~\cite{galar2011overview} in Figure~\ref{fig:f1score}. Concerning AffectNet, 
% FaceTopoNet 
our model exhibits higher F1 scores than BreG-NeXt50 in all the 8 expression classes except happy and neutral. AffectNet is skewed toward happy and neutral expressions. Despite this imbalance, 
% FaceTopoNet 
our pipeline shows better results in the minority expression classes which are more difficult to recognize. With regards to FER2013, 
% FaceTopoNet 
our pipeline achieves better F1 scores in 4 out of the 7 expressions. The exceptions are surprise, happy, and neutral expressions, in which BreG-NeXt50 performs slightly better.

% \begin{figure*}[t]
%     \begin{center}
%     \includegraphics[width=0.7\textwidth]{fig9-mergedVersion.pdf}
%     \end{center}
% \caption{Prediction samples of FaceTopoNet on AffectNet and FER2013. Ground truth of each image is mentioned below each sample. The color \textcolor{green}{green} indicates correct classification, while the color \textcolor{red}{red} indicated misclassification.}
% \label{fig:samples}
% \end{figure*}

\begin{figure}[t]
    \begin{center}
    \includegraphics[width=0.83\linewidth]{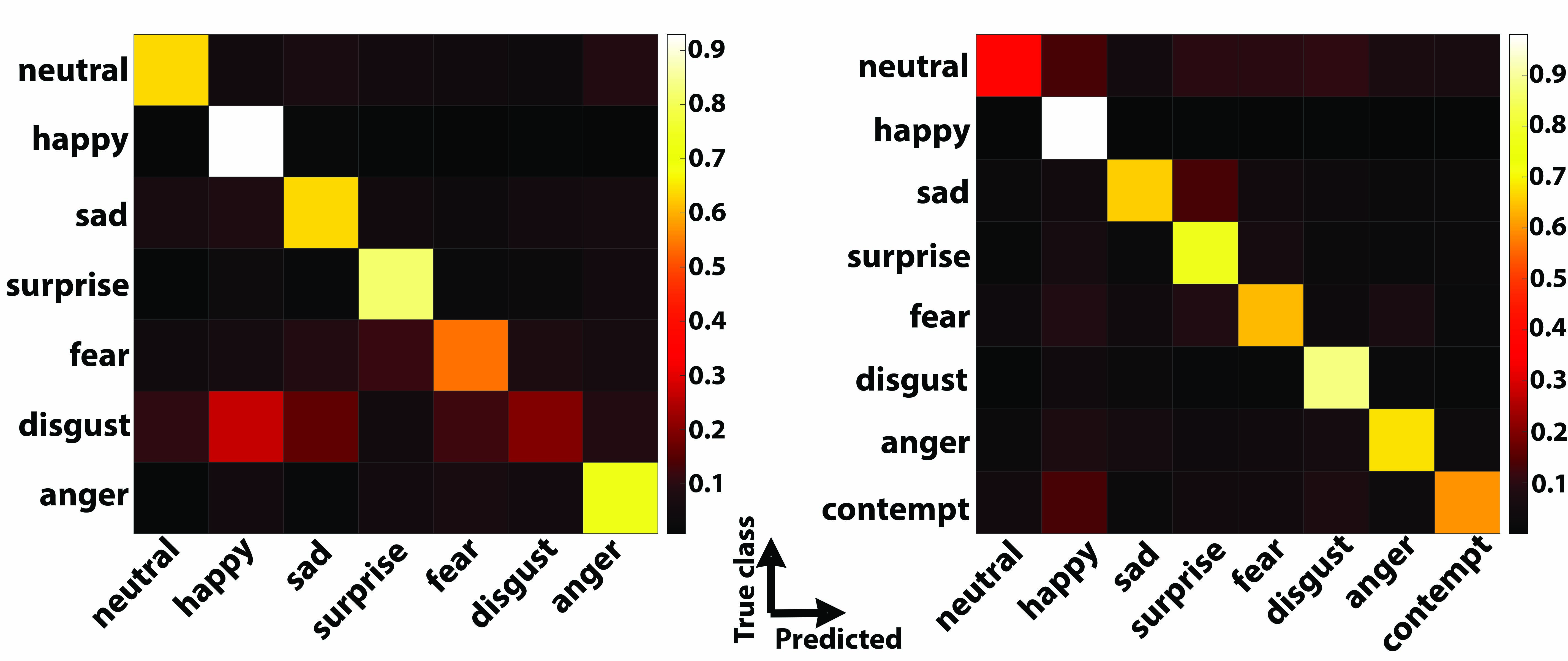}
    \end{center}
% \caption{Confusion matrices obtained by FaceTopoNet.}
\caption{Confusion matrices obtained by our model.}
\label{fig:ConfMat}
\end{figure}

Figure~\ref{fig:ConfMat} depicts the confusion matrices obtained by our method. For AffectNet, the largest confusions occur for contempt-happy (where contempt is the true class and happy is the predicted class), sad-surprise, and neutral-happy cases. Confusing contempt with happy is common in FER systems, since it is very hard to distinguish between these two even for humans~\cite{hasani2020breg}. In terms of FER2013, the largest confusion occurs in the disgust-happy case. The reason is that disgust and happy are the minority and majority classes respectively. Although we use focal loss to tackle imbalanced data, 
% FaceTopoNet 
some of the minority expressions were still classified as the majority classes. Other than disgust, 
% FaceTopoNet 
our pipeline shows good performance in recognizing true classes in FER2013 dataset.

% We visualize some of the correct and incorrect classification samples along with the corresponding output probabilities in Figure~\ref{fig:samples}. It can be observed that in the case of misclassified samples, the correct class labels always appear as the second highest output probability, indicating low rank-2 error. 

\subsection{Tree Topology Learning}

\begin{figure}[t]
    \begin{center}
    \includegraphics[width=0.88\columnwidth]{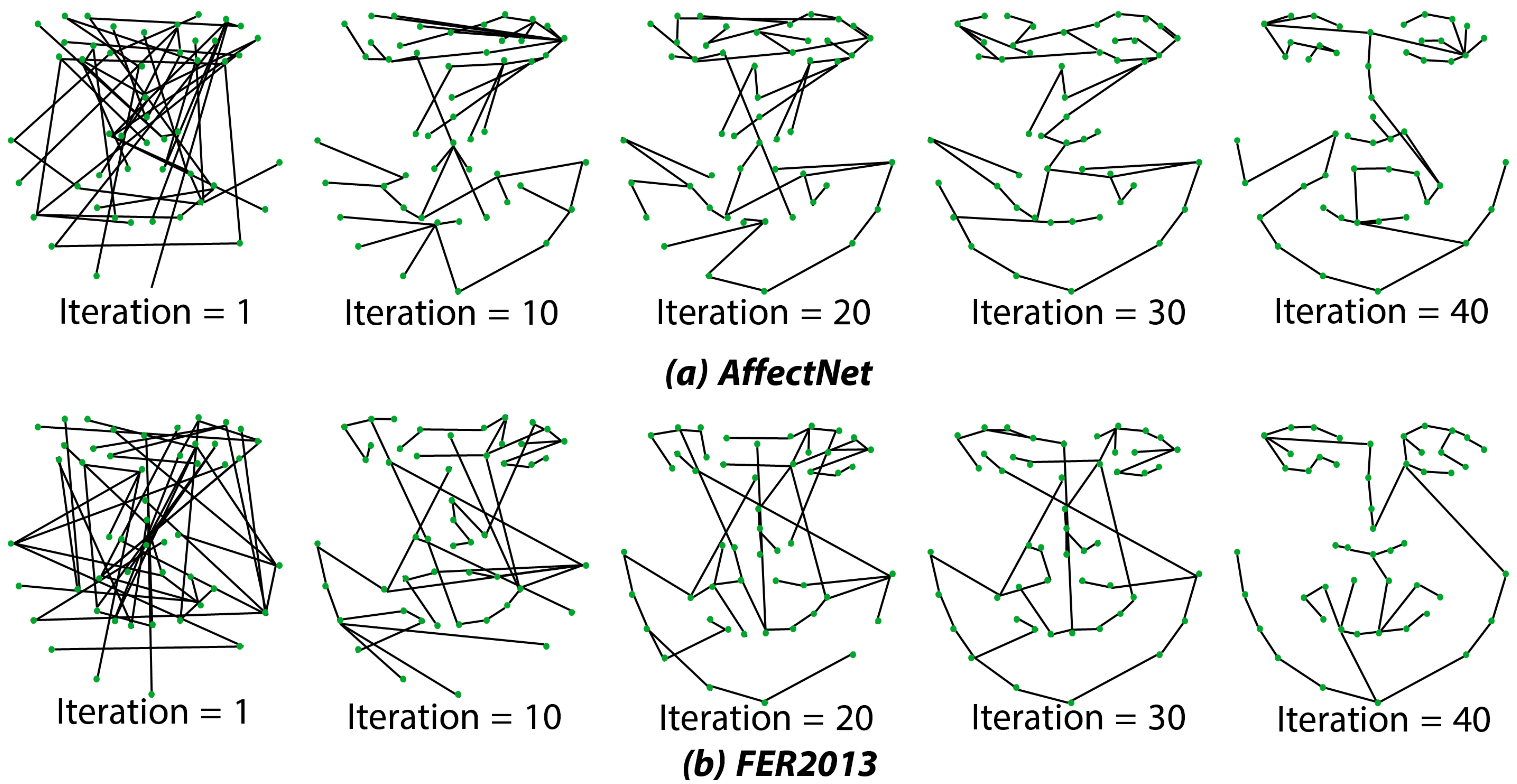}
    \end{center}
\caption{Evolution of face trees during the training phase.}
\label{fig:treeEvol}
\end{figure}

The topology evolution of the face trees during can provide insight into the learning process. Figure~\ref{fig:treeEvol} illustrates the convergence of the minimum-cost spanning trees obtained by our method, for AffetNet and FER2013 datasets in iterations 1, 10, 30, and 40. In the first iteration, 
% FaceTopoNet 
a random tree is generated by assigning random weights to the edges of the $k_{50}$. We observe that as training is being performed, face trees tend to resemble the structure of human faces more and more. As an obvious example, face trees formed in iteration 40 for both AffectNet and FER2013 dataset closely resemble human faces around the jaw line, mouth, nose, and eyes, while still being customized for each dataset. Furthermore, in iteration 40, large variations between the two face trees of AffectNet and FER2013 occurs near the mouth and eye regions. This is not surprising, since these regions are highly informative in FER, and therefore these regions are customized for each dataset.

% \begin{figure*}[t]
%     \begin{center}
%     \includegraphics[width=0.67\textwidth]{fig3j.pdf}
%     \end{center}
% \caption{Evolution of face trees during the training of FaceTopoNet.}
% \label{fig:treeEvol}
% \end{figure*}

\begin{table}[t]
\centering
\caption{Evaluation metrics on AffectNet and FER2013.}
\label{table:results}
\begin{tabular}{l|l|l|l} 
\hline
\textbf{Dataset}           & \textbf{Authors}                      & \textbf{Method} & \textbf{RR}     \\ 
\hline
\multirow{8}{*}{AffectNet} & Zeng et al.~\cite{zeng2018facialincon}          & CNN             & 57.31           \\
                           & Hewitt  et al.~\cite{hewitt2018cnn}       & CNN             & 58              \\
                           & Hua et al.~\cite{hua2019hero}           & Ensemble        & 62.11           \\
                           & Chen et al.~\cite{chen2019facial}          & Facial mask     & 61.50           \\
                           & Kollias et al.~\cite{kollias2020deep}       & Augmentation    & 60              \\
                           & Hasani et al.~\cite{hasani2020breg}        & BreG-NeXt32     & 66.74           \\
                           & Hasani et al.~\cite{hasani2020breg}        & BreG-NeXt50     & 68.50           \\
                           & \textbf{Ours}                         & Tree learning     & \textbf{70.02}  \\ 
\hline
\multirow{6}{*}{FER2013}   & Tang et al.~\cite{tang2013deep}          & CNN             & 69.3            \\
                           & Mollahosseini et al.~\cite{mollahosseini2016going} & CNN             & 66.4            \\
                           & Vielzeuf et al.~\cite{vielzeuf2017temporal}      & CNN             & 71.2            \\
                           & Hasani et al.~\cite{hasani2020breg}        & BreG-NeXt32     & 69.11           \\
                           & Hasani et al.~\cite{hasani2020breg}        & BreG-NeXt50     & 71.53           \\
                           & \textbf{Ours}                         & Tree learning     & \textbf{72.66}  \\
\hline
\end{tabular}
\end{table}

\begin{table}[t!]
\begin{center}
\caption{Ablation experiments.}
\label{table:abblation}
\footnotesize
\begin{tabular}{c|c|c|c|c}
\hline
\textbf{Removed} & \multicolumn{2}{|c|}{\textbf{RR}} &  \multicolumn{2}{|c}{\textbf{Drop in RR}} \\
\cline{2-5}
 & \textbf{AffectNet} & \textbf{FER2013} & \textbf{AffectNet} & \textbf{FER2013}\\
 \hline
 \hline
Tree topology & 66.51\% & 67.11\% & 5.0\% & 7.60\%\\
Structure stream & 64.90\% & 66.02\% & 7.30\% & 9.14\%\\
Texture stream & 62.14\% & 63.17\% & 11.25\% & 13.06\%\\
Fusion strategy & 66.87\% & 68.24\% & 4.50\% & 6.08\%\\
\hline
\end{tabular}
\end{center}
\end{table}

\subsection{Ablation Study}
To illustrate the contribution of each component of 
% FaceTopoNet, 
the proposed pipeline, we perform ablation experiments by systematically removing individual components of the model. Table~\ref{table:abblation} compares the recognition rates obtained by the following models: (\textit{a}) 
% FaceTopoNet 
the model without tree topology learning where we adopt a random tree to form the input embeddings; (\textit{b}) 
% FaceTopoNet 
the model without the structure stream, in which the coordinates of facial landmarks along with the learned tree topology are utilized to form the embeddings; (\textit{c}) 
% FaceTopoNet 
the model without structure stream where the output embedding of ResNet50 along with the learned tree topology is used to form the embeddings; and (\textit{d}) 
% FaceTopoNet 
the model with a simple concatenation of the output embeddings from the structure and texture streams, instead of using the 2-stream fusion strategy. 
% Results suggests that \textit{(1)} Tree topology learning is an effective component in FaceTopoNet, the removing of which results in a significant performance drops of 5.0\% and 7.60\% in AffectNet and FER2013 datasets respectively \textit{(2)} The contribution of texture stream is higher than that of the structure stream, which is not surprising. The reason lies in the fact that texture stream uses pretrained ResNet 50 which has over 23 million parameters. \textit{(3)} Although 2-fusion strategy uses attention mechanism, its relative contribution is lower than that of the tree topology learning. This demonstrates the importance of the tree topology learning in the pipeline. 
The results presented in Table~\ref{table:abblation} reveal that: (1) tree topology learning is an effective component in 
% FaceTopoNet 
our proposed pipeline whose removal results in significant performance drops of 5.0\% and 7.60\%, respectively for AffectNet and FER2013 datasets; (2) the contribution of the texture stream is higher than that of the structure stream due to the exploitation of the pre-trained ResNet50 in the texture stream; (3) relative contribution of the employed fusion strategy is lower than that of the tree topology learning. This demonstrates the importance of the tree topology learning in the pipeline.

\section{Conclusion}

In this paper we propose a novel end-to-end method called deep face topology network for facial expression recognition. Our model assigns weights to the edges of a complete graph representing all the possible connections between extracted facial landmarks and integrates them into an optimization process. These weights along with the complete graph are used to generate a minimum cost spanning tree (MST), whose traversal generates an optimal sequence. This sequence is used to form an embedding to feed an LSTM network. Our solution has been integrated in an end-to-end two-stream pipeline. The first stream uses the spatial information of facial landmarks, while the other uses the texture information of patches around the landmarks. A fusion strategy has finally been adopted to combine the two. Experiments have been performed with two large-scale AffectNet and FER2013 datasets, where the results show the superiority of our method in comparison to prior work. The main limitation of this work is the optimization speed, which we will address in future works by using policy gradient methods and applying faster algorithms for finding the MST.

%%%%%%%%%%%%%%%%%%%%%%%%%%%%%%%%%%%%%%%%%%%%%%%%%%%%%%%%%%%%%%%%%%%%%%%%%%%%%%%%

{\small
\bibliographystyle{ieee}
\bibliography{egbib}
}

\end{document}